\documentclass[runningheads]{asmda2005}

\usepackage{graphicx}
\usepackage{asmda2005References} 
\usepackage{breakcites}
\usepackage{epsfig}
\usepackage{hhline}

\def\RR{{\rm I\hspace{-0.50ex}R}}

\def\BR{{\sc n'Roger}}
\def\R{{\sc Roger}}

\begin{document}
%
\title*{Preference Learning in Terminology Extraction: A ROC-based approach}
%
\toctitle{Preference Learning in Terminology Extraction: A ROC-based approach} 
%
\titlerunning{Preference Learning in Terminology Extraction: A ROC-based approach}
%
\author{
  J\'er\^ome Az\'e
  \and 
  Mathieu Roche
  \and
  Yves Kodratoff
  \and
  Mich\`ele Sebag
}
%
\index{Az\'e, J.}
\index{Mathieu, R.}
\index{Kodratoff, Y.}
\index{Sebag, M.}
%
\authorrunning{Az\'e et al.}
%
\institute{
  LRI -- Laboratoire de Recherche en Informatique\\
  UMR8623, CNRS, Universit\'e Paris Sud, 91405 Orsay Cedex, France\\
  (e-mail: {\tt \{aze,roche,yk,sebag\}@lri.fr})
}

\maketitle             

\begin{abstract}
A key data preparation step in Text Mining, Term Extraction selects the terms, 
or collocation of words, attached to specific concepts.
In this paper, the task of extracting relevant collocations is achieved 
through a supervised learning algorithm, exploiting a few collocations manually 
labelled as relevant/irrelevant. The candidate terms are described 
along 13 standard statistical criteria measures. 
From these examples, an evolutionary learning algorithm termed \R, based on the optimization 
of the Area under the ROC curve criterion, extracts an order on the candidate terms. 
The robustness of the approach is demonstrated on two real-world domain applications, 
considering different domains (biology and human resources) and different languages
(English and French).
\end{abstract}

\section{Introduction}

Besides the known difficulties of Data Mining, Text Mining presents specific 
difficulties due to the structure of natural language. In particular, the polysemy and 
synonymy effects are dealt with by constructing ontologies or terminologies 
\cite{bourigault99term}, structuring the words and their meanings in the domain 
application. 
A preliminary step for ontology construction is to extract the terms, or word collocations, 
attached to the concepts defined by the expert \cite{bourigault99term,Xtract}. Term Extraction
actually involves two steps: the detection of the relevant collocations, and their 
classification according to the concepts. 

This paper focuses on the detection of relevant collocations, and presents a learning algorithm 
for ranking collocations with respect to their relevance, in the spirit of \cite{Cohen98}.
An evolutionary algorithm termed \R, based on the optimization of the Receiver Operating
Characteristics (ROC) curve  \cite{Ferri02,RossetICML04}, and already described in 
previous works \cite{ICDM03,EA03}, is applied to a few collocations manually labelled as
relevant/irrelevant by the expert. The optimization of the ROC curve is directly related to the 
recall-precision tradeoff in Term Extraction (TE). 

The paper is organized as follows. Section \ref{termExtraction} briefly reviews the main 
criteria used in TE. Section \ref{overview} presents the \R\ (ROc-based 
GEnetic learneR) algorithm for the sake of self-containedness, and describes the bagging 
of the diverse hypotheses constructed along independent runs. 
Sections \ref{setting} et \ref{experiements} report on the experimental validation of the approach on 
two real-world domain applications, and  
the paper ends with some perspectives for further research.

\section{Measures for Term Extraction} \label{termExtraction}\label{state}

The choice of a quality measure among the great many criteria used in Text Mining
(see e.g., \cite{daille98,Feiyu2002,rocheIIS04})
is currently viewed as a decision making process; the expert has to find the criterion most 
suited to his/her corpus and goals. The criteria considered in the rest of the paper are:
{\small
\begin{itemize}
\item Mutual Information ($MI$) \cite{Church90} 
\item Mutual Information with cube ($MI^3$) \cite{daille98}  
\item Dice Coefficient ($Dice$) \cite{Smadja96} 
\item Log-likelihood ($L$) \cite{dunning93accurate} 
\item Number of occurrences + Log-likelihood ($Occ_L$)\footnote{$Occ_L$ is defined by ranking collocations according to their number of occurrences, and breaking the ties based on the term Log-likelihood.} \cite{roche_ecai2004} 
\item Association Measure ($Ass$) \cite{jacq97} 
\item Sebag-Schoenauer ($SeSc$) \cite{sebag88}
\item J-measure ($J$) \cite{Goodman88}  
\item Conviction ($Conv$) \cite{brin97beyond} 
\item Least contradiction ($LC$) \cite{azeRNTI04} 
\item Cote multiplier ($CM$) \cite{Lallich-2004} 
\item Khi2 test used in text mining ($Khi2$) \cite{manning99} 
\item T-test used in text mining ($Ttest$) \cite{manning99} 
\end{itemize}           
}

Vivaldi et al. \cite{vivaldi01} have shown that the search for a quality measure can be 
formalized as a supervised learning problem. Considering a training set, where each candidate
term is described from its value for a set of statistical criteria and labelled by the expert, 
they used Adaboost \cite{Schapire99} to automatically construct a classifier.

The approach presented in next section mostly differs from \cite{vivaldi01} as it learns 
an ordering function (term $t_1$ is more relevant than term $t_2$) instead of a boolean 
function (term $t$ is relevant/irrelevant).

\section{Learning ranking functions} \label{overview}
This section first briefly recalls the \R\ ({\em ROc-based GEnetic learneR}) algorithm, 
used for learning a ranking hypothesis and 
first described in \cite{EA03,ICDM03}. The \BR\ variant used in this paper involves two 
extensions: i) the use of non-linear ranking hypotheses; ii) the exploitation of the ensemble of hypotheses learned along independent runs of \R.
Using the standard  notations, the dataset  $\cal E$ $= \{({\bf x}_i,y_i), i = 1..n, 
{\bf x}_i \in \RR^d, y_i \in \{-1,+1\}\}$ includes $n$ collocation examples, where each collocation ${\bf x}_i$
 is described by the value of $d$ statistical criteria, and its label $y_i$ denotes whether
collocation ${\bf x}_i$ is relevant.

\subsection{\R}
The learning criterion used in \R\ is the Wilcoxon rank test, measuring the probability
that a hypothesis $h$ ranks ${\bf x}_i$ higher than ${\bf x}_j$ when ${\bf x}_i$ is a positive
and ${\bf x}_j$ is a negative example:
\begin{equation} {\cal W}(h) = Pr (h(x_i) > h(x_j)~|~ y_i > y_j) \label{W}\end{equation}
This criterion, with quadratic complexity in the number $n$ of examples\footnote{Actually, the 
computational complexity is in ${\cal O}(n \log n)$ since ${\cal W}(h)$ is proportional to the 
sum of ranks of the positive examples.} offers an increased stability compared to the 
misclassification rate ($Pr (h(x_i).y_i > 0)$, with linear complexity in $n$); see 
\cite{RossetICML04} and references therein. The Wilcoxon rank test is equivalent to the area 
under the ROC (Receiver Operating Characteristics) curve \cite{Yan-ICML03}. 
This curve, intensively used in 
medical data analysis, shows the trade-off between the true positive rate (the fraction of 
positive examples that are correctly classified, aka recall) and the false positive rate (the 
fraction of negative examples that are misclassified) achieved by a given hypothesis/classifier/learning algorithm. Therefore, the area under the ROC curve (AUC) does not depend on 
the imbalance of the training set  \cite{Kolcz03}, as opposed to other measures 
such as Fscore \cite{Caruana04}. The ROC curve also shows the misclassification rates achieved 
depending on the error cost coefficients \cite{Domingos99}. 
For these reasons, \cite{Bradley97} argues  
the comparison of the 
ROC curves attached to two learning algorithms to be more fair and informative, than comparing 
their misclassification rates only. Accordingly, the area under the ROC curve defines a 
new learning criterion, used e.g. for the evolutionary optimization of neural nets 
\cite{Fogel}, or the greedy search of decision trees \cite{Ferri02}.

In an earlier step \cite{EA03}, the search space $\cal H$ considered is that of linear 
hypotheses (${\cal H} = \RR^d$). To each vector $w$ in $\RR^d$ is attached hypothesis
$h_w$ with $h_w(x) = <w,x>$, where $<w,x>$ denotes the scalar product of $w$ and $x$. 
Hypothesis $h$ defines an order on $\RR^d$, which is evaluated from 
the Wilcoxon rank test on the training set ${\cal E}$ (Eq. \ref{W}), measured after
  cross-validation.

The combinatorial optimization problem defined by Eq. \ref{W}, thus mapped onto a 
numerical optimization problem, is tackled by Evolution Strategies (ES). ES are the 
Evolutionary Computation algorithms that are best suited to parameter optimization; the 
interested reader is referred to \cite{Baeck-book} for an extensive presentation. 
In the rest of the paper, \R\ employs a $(\mu+\lambda)$-ES, involving the generation of 
$\lambda$ offspring from $\mu$ parents through uniform crossover and self-adaptive 
mutation, and 
deterministically selecting the next $\mu$ parents from the best $\mu$ parents $+$ $\lambda$ offspring. 
 
\subsection{Extensions}
An extension first presented in \cite{ECML04} concerns the use of non-linear hypotheses. 
Exploiting the flexibility of Evolutionary Computation, the search space ${\cal H}$ is
set to $\RR^d \times \RR^d$; each hypothesis $h = (w,c)$, composed of a weight vector $w$ and 
a center $c$, associates to $x$ the weighted $L_1$-distance of $x$ and $c$:

$$h(x = (x_1,...,x_d)) = \sum_{i=1}^d w_i |x_i - c_i|$$
It must be noted that this representation allows \R\ for searching (a limited kind of) 
non linear hypotheses, by (only) doubling the size of the linear search space. Previous 
work has shown that non-linear \R\ significantly outperforms linear \R\ for some text mining applications \cite{roche_ecai2004}.

A new extension, inspired from ensemble learning \cite{Breiman98}, exploits the 
hypotheses $h_1,\ldots,h_T$ learned along $T$ independent runs of \R.
The aggregation of the (normalised) $h_i$, referred to as $H$, associates to each example $x$ the 
median value of $\{h_1(x), \ldots,h_T(x) \}$.

\section{Goals of Experiments and Experimental Setting}\label{setting}
The goal of experiments is twofold. On one hand, the ranking efficiency of \BR\ will be 
assessed and compared to that of state-of-the-art supervised learning algorithms, specifically
Support Vector Machines with linear, quadratic and Gaussian kernels, using SVMTorch
implementation\footnote{http://www.idiap.ch/machine\_learning.php?content=Torch/en\_OldSVMTorch.txt} with default options. Due to space
limitations, only  ensemble-based non-linear \R, termed \BR, will be 
considered.

On the other hand, the results provided by \BR\ will be interpreted and 
discussed with respect to their intelligibility.
The experimental setting is as follows. An experiment is a 5-fold stratified 
cross-validation process;
on each fold, i) SVM learns a hypothesis $h_{SVM}$; ii) \R\ is launched 21 times, and the 
bagging of the 21 learned hypotheses constitutes the hypothesis $h_{\BR}$ learned by \BR;
iii) both hypotheses are evaluated on the fold test set and the associated 
ROC curve (True Positive Rate {\em vs} False Positive Rate) is constructed.
The AUC curves are averaged over the 5 folds.

The overall results reported in the next section are averaged over 10 experiments 
(10 different splits of the dataset into 5 folds).

The \R\ parameters are as follows: $\mu = 20; \lambda = 100$; the self adaptative mutation 
rate is 1.; the uniform crossover rate is $.6$.

\section{Empirical validation} \label{experiements}
After describing the datasets, this section reports on the comparative performances 
of the algorithms, and inspects the results actually provided by \BR.

\subsection{Datasets}
In both domains, the data preparation step  \cite{rocheIIS04} allows for categorizing the 
word collocations depending on the grammatical tag of the words (e.g. Adjective, Noun).
 
A first corpus related to Molecular Biology involves 6119 paper abstracts in English
(9,4 Mo) gathered from queries on Medline\footnote{http://www.ncbi.nlm.nih.gov/entrez/query.fcgi}. 
The 1028 Noun-Noun collocations occurring more than 4 times are labelled by the expert; 
the dataset includes a huge majority of relevant collocations (Table \ref{tab:collocations_expertisees}).

A second corpus related to Curriculum Vitae\footnote{Courtesy of the VediorBis Foundation.} involves 582 CVs in French (952 Ko). The ``Frequent CV'' dataset includes the 376 Noun-Adjective collocations with at least 3 occurrences (two hours labelling required), with a 
huge majority of relevant collocations. The ``Infrequent CV'' dataset includes the 2822 Noun-Adjective 
collocations occurring once or twice (two days labelling required), with a significantly 
different distribution of relevant/irrelevant collocations (Table \ref{tab:collocations_expertisees}).
Examples of relevant {\em vs} irrelevant collocations are respectively {\em comp\'etences informatiques}
and {\em euros annuels}; 

although both collocations make sense, only the first one conveys useful
information for the management of human resources.

\begin{table}
\centering
        \begin{minipage}[!]{11cm}
\begin{scriptsize}
\centering
      \begin{tabular}[!]{|c|c|c|c|} 
        \hline
        {\bf Collocations}  & {\bf \# collocations} & {\bf Relevant} & {\bf Irrelevant} \\
        \hline
        Molecular Biology & 1028  & 90.9\% & 9.1\% \\
        \hline
        CV, Frequent collocations  & 376   & 85.7\%  & 14.3\% \\
        \hline
        CV, Infrequent collocations &  2822 & 56.6\%  & 43.4\% \\
        \hline
      \end{tabular}
      \caption{Relevant and irrelevant collocations.} \label{tab:collocations_expertisees}
\end{scriptsize}
\end{minipage}
\end{table}
\vspace{-0.5cm}

\subsection{Ranking accuracy}

After the experimental setting described in section \ref{setting}, 
Table  \ref{tab:AUC_Roger_SVM} compares the average AUC achieved for \BR\ and SVMTorch with 
 linear, Gaussian and quadratic kernels.
On these domain applications, both supervised learning approaches significantly improve
on the statistical criteria standalone (Table  \ref{tab:AUCmesures}). Further, 
\BR\ improves significantly on SVM using any kernel, excepted on 
the {\em Infrequent CV} dataset. A tentative
interpretation for this result is based on the fact that this dataset is the most balanced
one; SVM has some difficulties to cope with imbalanced datasets.

\begin{table}[htbp]
        \begin{minipage}[!]{11cm}
  \begin{scriptsize}
\centering

\begin{tabular}[!]{|c|c|c|c|c|} 
        \hline
        {\bf Corpus}   & \BR\ & \multicolumn{3}{c|}{{\sc SVM} ($\sim$ 1.5s/fold)} \\
        & ($\sim$ 17s/fold)    & Linear & Gaussian & Quadratic \\
        \hline
        \hline
        {\bf Molecular Biology (MB)}       & $0.73 \pm 0.05$ & $0.50 \pm 0.08$ & $0.46 \pm 0.08$ & $0.59 \pm 0.08$ \\ 
        \hline
        {\bf Frequent CV (F-CV)}  & $0.64 \pm 0.08$ & $0.48 \pm 0.08$ & $0.48 \pm 0.08$ & $0.50 \pm 0.10$ \\
        \hline
        {\bf Infrequent CV (I-CV)}      & $0.73 \pm 0.01$ & $0.72 \pm 0.01$ & $0.72 \pm 0.02$ & $0.71 \pm 0.02$\\
        \hline
      \end{tabular}
      \caption{Ranking accuracy (Area under the ROC curve) of learning algorithms.} \label{tab:AUC_Roger_SVM}

      \begin{tabular}[!]{|c|c|c|c|c|c|c|c|c|c|c|c|c|c|} 
        \hline
        {\bf Corpus}   & $MI$ & $MI^3$ & $Dice$ & $L$ & $Occ_L$ & $Ass$ & $J$ & $Conv$ & $SeSc$ & $CM$ & $LC$ & $Ttest$ & $Khi2$ \\
        \hline
        \hline
        {\bf MB}  & $0.30$ & $0.35$ & $0.31$ & $0.42$ & $0.57$ & $0.31$ &  {\bf 0.59} & $0.35$ &  $0.43$ & $0.31$ & $0.46$ &  $0.31$ &  $0.31$ \\ 
        \hline
        {\bf F-CV}       & $0.31$ & $0.40$ & $0.39$ & $0.43$ & {\bf 0.58} & $0.32$ &  {\bf 0.58} & $0.39$   & $0.40$   & $0.31$ & $0.44$ &  $0.36$ & $0.36$ \\ 
        \hline
        {\bf I-CV}    & $0.29$&  $0.30$&  $0.33$ & $0.30$ &$0.37$ &$0.29$& {\bf 0.50} & $0.40$ & $0.39$& $0.30$ & $0.45$ & $0.30$ & $0.30$\\
        \hline
      \end{tabular}
     \caption{Ranking accuracy (Area under the ROC curve) of statistical criteria.} \label{tab:AUCmesures}
\end{scriptsize}
\end{minipage}
\end{table}

\begin{figure}[htbp]
        \begin{minipage}[!]{11cm}
\centering
      \epsfig{file=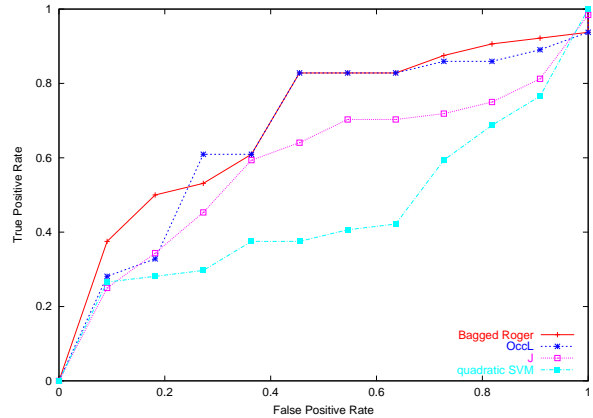, angle=-90,width=8cm}
    \caption{ROC Curves on Frequent Collocations of CV corpus (for the test set of $RF$).}
\label{fig:ROC_Test} 
\end{minipage}
\end{figure}

A more detailed picture is provided by Fig. \ref{fig:ROC_Test}, showing the 
ROC curve associated to SVM, \BR\ and the $Occ_L$ and $J$ measures on the {\em Frequent CV} dataset
on a representative fold (termed $RF$ in this paper).
Interestingly, the major differences between \BR\ and the other
measures are seen at the beginning of the curve, i.e. they concern the top ranked collocations.
Typically, a recall (True Positive Rate) of 50\% is obtained for 
18\% false positive with \BR, against 23\% with $Occ_L$, 31\% with $J$ measures 
and 68\% for quadratic SVM\footnote{SVM ROC Curves is not significant as its AUC is lower than .5 on this test fold.}.

In summary, \BR\ improves the accuracy of the top-ranked collocations, and therefore the 
satisfaction and productivity of the expert if he/she only examines the top results.
A proof of principle of the generality of the approach has been presented in 
 \cite{rocheIIS04}, as the ranking function learned from one corpus, in one language, 
was found to outperform the standard statistical criteria when applied on the 
other corpus, in another language.

\vspace{-0.5cm}

\subsection{Analysis of a ranking function}

As shown in \cite{ECML04}, the weights associated to distinct features by \R\ can provide
some insights into the relevance of the features. Accordingly, the hypotheses constructed
by \BR\ are examined.
\begin{figure}[htbp]
        \begin{minipage}[!]{11cm}
\centering
\epsfig{file=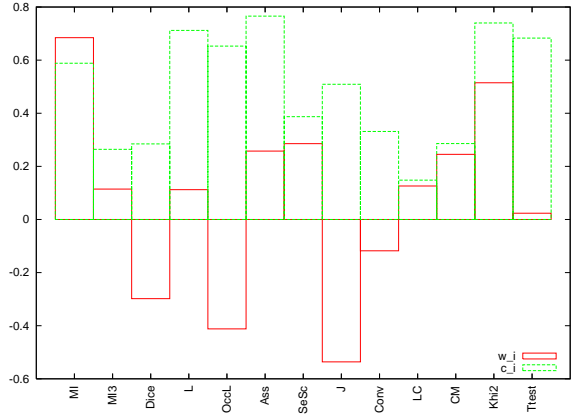, angle=-90,width=8cm}

\caption{Weights ($w_j, c_j$) on the frequent CVs (for the learn set of $RF$).}
    \label{fig:weight_function_5.1}
\end{minipage}
\end{figure}

Fig. \ref{fig:weight_function_5.1} displays the weights and 
center coordinates of all 13 features (section \ref{state}) for a representative 
\R\ hypothesis $h$ (closest to the ensemble \BR\ hypothesis $H$) 
learned on a fold of the {\em Frequent
CV} dataset. Although AUC$(h)$ is lower than that of $H$ (.61 {\em vs} .64), 
it still outpasses that of standalone features (statistical criteria). 

As could have been expected, \R\ detects that the  mutual 
information ($MI$) criterion does badly (AUC($MI$)= .31, Table \ref{tab:AUCmesures}),
with a high center $c_{MI}$ and weight $w_{MI}$ values (collocations with high $MI$ are less relevant, 
everything else being equal). 
Inversely, as the $Occ_L$ criterion does well (AUC($Occ_L$) = .58), the center $c_{Occ_L}$ is high associated with a highly negative weight $w_{Occ_L}$
(collocations with low $Occ_L$ are less relevant, everything else being equal) (see Tab. \ref{tab:ExamplesMeasuresWeights}).

Although these tendencies could have been exploited by linear hypotheses, this is no 
longer the case for the $J$ criterion (AUC($J$) = .58): interestingly, the center $c_{J}$ takes
on a medium value, with a high negative weight $w_{J}$. This is interpreted as collocations with either
too low {\em or too high} values of $J$, are less relevant everything else being equal.
The current limitation of the approach is to provide a ``conjunctive'' description 
of the region of relevant collocations\footnote{In the sense that a single center $c$ is considered, 
though the condition {\em far from $c_i$} actually corresponds to a disjunction.}. 

\vspace{-0.5cm}
\begin{table}
\begin{scriptsize}
\centering
      \begin{tabular}[!]{|c|c|c|c|c|c|c|} 
        \hhline{~~|-|~|-|~|-}
        \multicolumn{1}{c}{} & & \multicolumn{1}{|c|}{{\bf $MI$ }} & & \multicolumn{1}{|c|}{{\bf $Occ_L$ }} & & \multicolumn{1}{|c|}{}\\
\multicolumn{1}{c}{} & & \multicolumn{1}{|c|}{{\bf $w_{MI}$ = 0.68}} & & \multicolumn{1}{|c|}{{\bf $w_{Occ_L}$ = -0.41}} &  & \multicolumn{1}{|c|}{\BR} \\
\multicolumn{1}{c}{} & & \multicolumn{1}{|c|}{{\bf $c_{MI}$ = 0.59}} &  & \multicolumn{1}{|c|}{{\bf $c_{Occ_L}$ = 0.65}} & &\\

        \hhline{-|~|-|~|-|~|-}
         {\bf Collocation}  & & {\bf Rank} &  & {\bf Rank} &  & {\bf Rank}\\
        \hhline{-|~|-|~|-|~|-}
        
        exp\'erience commerciale & & 297 & & 258 &  & 1 \\
        formation informatique   & & 300 & & 123 &  & 2 \\
        soci\'et\'e informatique & & 298 & & 299 &  & 3 \\
        gestion informatique     & & 299 & & 76  &  & 4 \\
        
        \hhline{-|~|-|~|-|~|-}
        \hhline{-|~|-|~|-|~|-}

        colonne morris          & & 1 & & 211 & & 90 \\
        bouygue telecom         & & 2 & & 213 & & 298\\
        fromagerie riches-mont  & & 3 & & 212 & & 297\\
        sauveteur secouriste    & & 4 & & 151 & & 296\\

        \hhline{-|~|-|~|-|~|-}
        \hhline{-|~|-|~|-|~|-}
        
        exp\'erience professionelle & & 146 & & 1 & & 300\\
        ressource humaine           & & 44  & & 2 & & 299\\
        baccalaur\'eat professionel & & 193 & & 3 & & 22 \\
        baccalaur\'eat scientifique & & 148 & & 4 & & 58 \\

         \hhline{-|~|-|~|-|~|-}
      \end{tabular}
      \caption{Rank of relevant collocations given with 2 measures ($MI$ and $Occ_L$) and \BR. For each measure the weights ($w_i$, $c_i$) used by \BR\ are given (on the learn set of $RF$).} \label{tab:ExamplesMeasuresWeights}
\end{scriptsize}
\end{table}
\vspace{-1cm}

\section{Discussion and Perspectives}
The main claim of the paper is that supervised learning can significantly contribute to 
the Term Extraction task in Text Mining. 
Some empirical evidence supporting this claim have been presented, related to 
two corpora with different domain applications and languages. 
Based on a domain- and language-independent description of the 
collocations along a set of standard statistical 
criteria, and on a few collocations manually labelled as relevant/irrelevant by the expert, 
a ranking hypothesis is learned. 

The ranking learner \BR\ used in the experiments is based on the optimization 
of the combinatorial Wilcoxon rank test criterion, using an evolutionary computation 
algorithm. Two new features, the use of non-linear hypotheses and the exploitation
of the ensemble of hypotheses learned along independent runs of \R, have been 
exploited in \BR.

Further research is concerned with enriching the description of collocations, e.g. adding 
typography-related indications (e.g. distance to the closest typographic signs) or distance to the 
closest Noun, possibly providing 
additional cues on the role of relevant collocations. 
Another perspective is to extend \R\ using multi-modal and multi-objective 
evolutionary optimization \cite{Deb01}, e.g. enabling to characterize several types of 
relevant collocations in a single run.
A long-term goal is to study along a variety of domain applications and expert goals, 
the eventual regularities associated to i) the (domain and language independent) 
description of the relevant collocations; ii) the ranking hypotheses.

~\\
{\bf Acknowledgment:} We thank Oriane Matte-Tailliez for her expertise and labelling of the 
Molecular Biology dataset and Mary Felkin who did her best to improve the readability of this paper.
The authors are partially supported by the PASCAL Network of Excellence, IST-2002-506778.

\bibliographystyle{asmda2005References}

\bibliography{article-asmda2005.aze_al}

\end{document}